\documentclass{article}

\PassOptionsToPackage{numbers, sort&compress}{natbib}
\usepackage[dblblindworkshop,final]{neurips_2026}

\usepackage[utf8]{inputenc}
\usepackage[T1]{fontenc}
\usepackage{hyperref}
\usepackage{url}
\usepackage{booktabs}
\usepackage{amsmath}
\usepackage{amsfonts}
\usepackage{amssymb}
\usepackage{nicefrac}
\usepackage{microtype}
\usepackage{xcolor}
\usepackage{colortbl}
\usepackage{graphicx}
\usepackage{multirow}
\usepackage{tikz}
\usetikzlibrary{arrows.meta, positioning, fit, decorations.pathreplacing, calc}
\usepackage{cleveref}
\DeclareMathSizes{10}{10}{7.4}{6.2}

\definecolor{hlrow}{RGB}{255,252,204}
\definecolor{tokfill}{RGB}{250,238,205}
\definecolor{clsfill}{RGB}{243,213,220}

\crefname{table}{Table}{Tables}
\Crefname{table}{Table}{Tables}
\crefname{figure}{Figure}{Figures}
\Crefname{figure}{Figure}{Figures}
\crefname{section}{Section}{Sections}
\Crefname{section}{Section}{Sections}
\crefname{appendix}{Appendix}{Appendices}
\Crefname{appendix}{Appendix}{Appendices}
\crefname{equation}{Eq.}{Eqs.}
\Crefname{equation}{Eq.}{Eqs.}

\title{Prune-and-Merge: Training-Free Scaling of Vision Transformers\thanks{\href{https://github.com/Chalhotra/ViT-Token-Economy}{Official Repository}}}
\workshoptitle{GlobalSouthAI @ NeurIPS 2026}
\author{%
  Vatsal \\
  Department of Mathematics\\
  Indian Institute of Technology Roorkee\\
  \texttt{vatsal@ma.iitr.ac.in}
  \And
  Soumil Malhotra \\
  Department of Mathematics\\
  Indian Institute of Technology Roorkee\\
  \texttt{soumil\_m@ma.iitr.ac.in}
  \And
  Priyanshu Chahal \\
  Department of Mathematics\\
  Indian Institute of Technology Roorkee\\
  \texttt{priyanshu@ma.iitr.ac.in}
  \And
  Sanjeev Kumar \\
  Department of Mathematics\\
  Indian Institute of Technology Roorkee\\
  \texttt{sanjeev.kumar@ma.iitr.ac.in}
}

\begin{document}

\maketitle

\begin{abstract}
  Vision Transformers achieve strong performance but incur high computational costs from
  quadratic token interactions. Training-free token reduction methods improve efficiency,
  but their robustness remains underexplored. We propose Prune-and-Merge ViT, a
  training-free framework for off-the-shelf models. It combines class-attention-based
  pruning with similarity-based merging to reduce redundancy while preserving semantic
  information. To evaluate robustness we introduce two metrics. Relative Accuracy Drop (RAD)
  measures relative accuracy degradation under transformations, and Prediction Flip Rate
  (PFR) quantifies prediction consistency. On pretrained ViT and DeiT, Prune-and-Merge
  reduces computation competitively while maintaining robustness and better accuracy than
  existing token reduction methods. Our analysis suggests that hybrid token reduction does
  not introduce hallucination-like behavior and preserves prediction stability for common
  image transformations.
\end{abstract}

\section{Introduction}

Vision Transformers (ViTs) forgo convolutional inductive biases and treat an image as a
sequence of visual tokens~\citep{dosovitskiy2021vit}. Non-overlapping patches are projected into \(N\)
tokens and passed, with a learnable class ([CLS]) token, through Transformer blocks. The
token count is constant across layers, so cost scales quadratically, \(O(N^2)\).

To address this issue, token pruning removes redundant tokens and merging aggregates similar
ones to cut computation~\citep{bolya2023tome, liang2022evit}. Many such methods are
training-free and apply directly to pretrained models. While such approaches improve efficiency, they
may introduce instability under input transformations, potentially affecting prediction
consistency. Aggressive reduction at lower keep rates can also degrade performance, as fewer
tokens remain to capture essential visual information. This motivates a systematic analysis of
prediction stability under aggressive reduction.

\begin{figure}[t]
  \centering
  \resizebox{0.88\linewidth}{!}{% Overview of the Prune-and-Merge ViT architecture (TikZ reconstruction).
\begin{tikzpicture}[
    font=\sffamily,
    stage/.style={draw, rounded corners=3pt, line width=0.7pt, minimum height=7mm,
                  text width=30mm, inner xsep=1pt, align=center,
                  font=\sffamily\bfseries\small},
    tok/.style={draw, rounded corners=1.2pt, line width=0.4pt, fill=tokfill,
                minimum width=4.4mm, minimum height=4.2mm, inner sep=0.5pt,
                font=\sffamily\small},
    cls/.style={tok, fill=clsfill, minimum width=6.8mm},
    grp/.style={draw, line width=0.4pt, rounded corners=1pt, inner xsep=1.6mm, inner ysep=1.4mm},
    lbl/.style={font=\sffamily\small, align=center},
    flow/.style={-{Latex[length=1.7mm,width=1.5mm]}, line width=0.9pt},
    fine/.style={-{Latex[length=1.3mm,width=1.2mm]}, line width=0.35pt},
    brc/.style={decorate, decoration={brace, mirror, amplitude=3pt}, line width=0.35pt},
    node distance=0pt,
  ]

  % ---------- Top pipeline ----------
  \node[stage] (s0) at (0,0)      {Patch Tokens + CLS Token};
  \node[stage] (s1) at (4.84,0)   {Token Pruning};
  \node[stage] (s2) at (9.64,0)   {Token Merging};
  \node[stage] (s3) at (14.00,0)  {Output Tokens};

  \draw[flow] (s0) -- (s1);
  \draw[flow] (s1) -- (s2);
  \draw[flow] (s2) -- (s3);

  \node[font=\sffamily\bfseries\small, above=1.1mm of s1] {Stage 1};
  \node[font=\sffamily\bfseries\small, above=1.1mm of s2] {Stage 2};

  % ---------- Input token strip ----------
  \begin{scope}[shift={(-1.67,-1.02)}]
    \node[cls] (i0) at (0,0) {[CLS]};
    \node[tok, right=0.5mm of i0] (i1) {T1};
    \node[tok, right=0.5mm of i1] (i2) {T2};
    \node[tok, right=0.5mm of i2] (i3) {T3};
    \node[tok, right=0.5mm of i3] (i4) {T4};
    \node[tok, right=0.5mm of i4] (i5) {T5};
    \node[right=0.5mm of i5, font=\small] (idots) {$\cdots$};
    \node[tok, right=0.5mm of idots] (iN) {TN};
    \draw[brc] ([yshift=-1.2mm]i0.south west) -- ([yshift=-1.2mm]iN.south east)
      node[midway, below=1.8pt, lbl] {N Tokens};
  \end{scope}

  % ---------- Stage 1: pruning ----------
  \begin{scope}[shift={(3.17,-1.02)}]
    \node[cls] (p0) at (0,0) {[CLS]};
    \node[tok, right=0.5mm of p0] (p1) {T1};
    \node[tok, right=0.5mm of p1] (p2) {T2};
    \node[tok, right=0.5mm of p2] (p3) {T3};
    \node[tok, right=0.5mm of p3] (p4) {T4};
    \node[tok, right=0.5mm of p4] (p5) {T5};
    \node[right=0.5mm of p5, font=\small] (pdots) {$\cdots$};
    \node[tok, right=0.5mm of pdots] (pN) {TN};
  \end{scope}

  \begin{scope}[shift={(3.66,-2.30)}]
    \node[cls] (q0) at (0,0) {[CLS]};
    \node[tok, right=0.5mm of q0] (q1) {T1};
    \node[tok, right=0.5mm of q1] (q2) {T2};
    \node[tok, right=0.5mm of q2] (q3) {T4};
    \node[right=0.5mm of q3, font=\small] (qdots) {$\cdots$};
    \node[tok, right=0.5mm of qdots] (qF) {TF};
    \node[below=0.5mm of q1, lbl] {0};
    \node[below=0.5mm of q2, lbl] {1};
    \node[below=0.5mm of qF, lbl] {K};
    \draw[brc] ([yshift=-4.2mm]q0.south west) -- ([yshift=-4.2mm]qF.south east)
      node[midway, below=1.8pt, lbl] {K+1 Tokens};
  \end{scope}

  \draw[fine] (p3.south) -- (qF.north);
  \draw[fine] (p4.south) -- (qF.north);
  \draw[fine] (p5.south) -- (qF.north);
  \draw[fine] (pN.south) -- (qF.north);

  % ---------- Stage 2: merging ----------
  \begin{scope}[shift={(8.54,-1.28)}]
    \node[tok] (a0) at (0,0) {0};
    \node[tok, below=1.5mm of a0] (a2) {2};
    \node[tok, below=1.5mm of a2] (a4) {4};
    \node[below=0.8mm of a4, font=\small] (advd) {$\vdots$};

    \node[tok] (b1) at (2.2,0) {1};
    \node[tok, below=1.5mm of b1] (b3) {3};
    \node[tok, below=1.5mm of b3] (b5) {5};
    \node[below=0.8mm of b5, font=\small] (bdvd) {$\vdots$};

    \node[grp, fit=(a0)(a4)(advd)] (boxA) {};
    \node[grp, fit=(b1)(b5)(bdvd)] (boxB) {};

    \node[above=0.5mm of boxA, lbl] {Group 1};
    \node[above=0.5mm of boxB, lbl] {Group 2};
    \node[below=0.7mm of boxA, lbl] {Even-indexed\\tokens};
    \node[below=0.7mm of boxB, lbl] {Odd-indexed\\tokens};

    \draw[fine] (a0.east) -- (b3.west);
    \draw[fine] (a2.east) -- (b1.west);
    \draw[fine] (a4.east) -- (b5.west);
  \end{scope}

  % ---------- Output token strip ----------
  \begin{scope}[shift={(12.77,-1.52)}]
    \node[cls] (o0) at (0,0) {[CLS]};
    \node[tok, right=0.5mm of o0] (o1) {T1'};
    \node[tok, right=0.5mm of o1] (o2) {T2'};
    \node[tok, right=0.5mm of o2] (o3) {T3'};
    \node[right=0.5mm of o3, font=\small] (odots) {$\cdots$};
    \node[tok, right=0.5mm of odots] (oN) {TN'};
  \end{scope}

\end{tikzpicture}}
  \caption{Overview of the Prune-and-Merge ViT architecture. The framework processes tokens
  through a dual-stage reduction at layers \(l \in \{3, 6, 9\}\). Stage 1 (pruning) filters
  out background tokens based on class-attention scores \(k_p\), while Stage 2 (merging)
  aggregates redundant salient tokens using bipartite matching \(k_m\).}
  \label{fig:arch}
\end{figure}
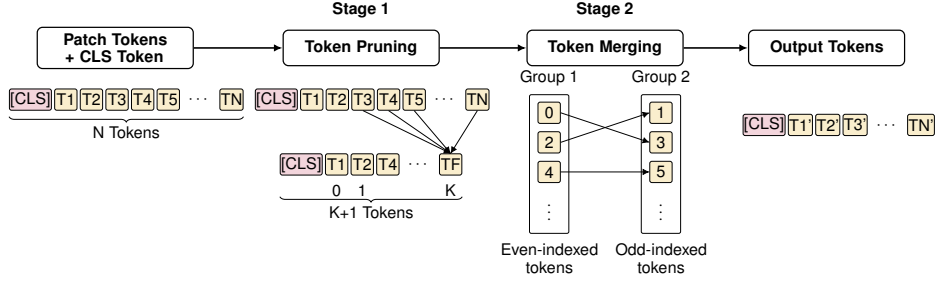

Existing approaches apply pruning or merging in isolation, which causes semantic loss or
hallucination-like errors under input transformations. We address this with Prune-and-Merge
ViT (\cref{fig:arch}), a plug-and-play, training-free framework that combines
class-attention-based pruning with similarity-driven merging for off-the-shelf
ViT~\citep{dosovitskiy2021vit} and DeiT~\citep{touvron2021deit}. The first stage discards
low-importance tokens via top-\(k\) selection on class-attention
scores~\citep{liang2022evit}, which filters noise while retaining salient features.
Subsequently, a merging stage aggregates the remaining redundant tokens by bipartite
matching on cosine
similarities~\citep{bolya2023tome}, preserving the semantic manifold in a compact bottleneck
\(\mathbf{T}' \in \mathbb{R}^{N' \times D}\), where \(N' \ll N\). Unlike single-strategy
methods, our framework controls the token budget finely: adjusting the pruning ratio and
the merging threshold preserves semantic stability.

To evaluate robustness we introduce two metrics: Relative Accuracy Drop (RAD), which
measures relative accuracy degradation under input transformations, and Prediction Flip
Rate (PFR), which quantifies prediction consistency between original and transformed
inputs. Experiments
on pretrained ViT and DeiT show that Prune-and-Merge ViT competitively reduces computation
while maintaining robustness and achieving strong accuracy. Our analysis further shows that
hybrid token reduction preserves prediction stability under common image transformations.

Our contributions are summarized as follows:
\begin{itemize}
  \item Prune-and-Merge ViT, a training-free hybrid token reduction framework.
  \item An analysis showing that hybrid pruning and merging preserve stability while reducing computation.
  \item RAD and PFR, which evaluate robustness and prediction stability under
  semantics-preserving transformations.
  
\end{itemize}

\section{Related work}
\label{sec:related}

ViT~\citep{dosovitskiy2021vit} and DeiT~\citep{touvron2021deit} keep the sequence length
fixed, so inference stays costly. DynamicViT~\citep{rao2021dynamicvit} learns a pruning
module, but it has to be trained. EViT~\citep{liang2022evit} and Top-K need no training.
They rank patches by class-attention and discard the rest, while
V-Pruner~\citep{yao2026vpruner} ranks globally rather than locally, and
\citet{zhan2024statespace} carry the paradigm to state-space models.

ToMe~\citep{bolya2023tome} fuses similar tokens by bipartite matching instead, which
shortens the sequence physically rather than masking it. The same mechanism accelerates
Stable Diffusion~\citep{bolya2023tomesd}, and \citet{haurum2023which} find that ToMe works
best when at least half the token budget is kept. Spatial-preserving
hybrids~\citep{mao2025efficient, erak2025adaptive} enhance locality and
VisionTrim~\citep{yu2026visiontrim} favors throughput, but neither line offers robustness
guarantees, and hybrids of the two families remain underexplored.

We address that gap in the plug-and-play spirit of ToCom~\citep{jie2024tocom} and its
vision-language extensions~\citep{zhang2025hybridtoken}, where transfer to a narrow subset
such as ImageNet-100 needs careful alignment~\citep{peng2025robustness};
see~\cref{app:related}.

\section{Method}
\label{sec:method}

\subsection{Framework overview}
\label{sec:overview}

We propose Prune-and-Merge ViT (see~\cref{fig:arch}), a dual-stage, training-free framework.
It treats token reduction as sequential filtering and compression. Given \(N\) input tokens
\(\mathbf{x}\), we apply a hierarchical reduction at layers \(l \in \{3, 6,
9\}\)~\citep{haurum2023which}. This schedule preserves high-resolution spatial information in
early layers before progressively reducing the sequence length in deeper blocks.

The composite keep rate \(K = k_p \cdot k_m\) governs reduction intensity at each stage. Here,
\(k_p \in (0, 1]\) is the pruning keep rate, the fraction of tokens retained according to
semantic importance; pruning thus acts as a semantic filter. Conversely, \(k_m \in (0, 1]\) is
the merging keep rate, the fraction remaining after bipartite matching of the pruned subset.
This multiplicative form allows fine-grained control over the total token budget while
ensuring the reduction scales proportionally with the remaining sequence length.

\subsection{Stage 1: class-attention pruning}
\label{sec:stage1}

Stage 1 retains semantically important tokens using the [CLS] token's class-attention,
following EViT~\citep{liang2022evit}. Multi-head self-attention over input tokens
\(\mathbf{x} \in \mathbb{R}^{B \times N \times C}\) produces an attention map
\(\mathbf{A} \in \mathbb{R}^{B \times H \times N \times N}\). We measure importance by [CLS]
attention to the \(P = N - n_s\) patch tokens,
\begin{equation}
  \label{eq:cls-attn}
  \mathbf{A}_{\mathrm{cls}} = \mathbf{A}[{:},{:},0,n_s{:}] \in \mathbb{R}^{B \times H \times P},
  \qquad
  \bar{\mathbf{a}} = \frac{1}{H} \sum_{h=1}^{H} \mathbf{A}_{\mathrm{cls}}^{(h)}
  \in \mathbb{R}^{B \times P},
\end{equation}
where \(n_s\) counts the special tokens. The top-\(k\) attentive tokens are selected
using \(k_p\),
\begin{equation}
  \label{eq:topk}
  k = \max\!\left(1, \lceil k_p \cdot P \rceil\right), \qquad
  \mathcal{I}_{\text{att}} = \mathrm{TopK}(\bar{\mathbf{a}}, k).
\end{equation}
To prevent total loss of background context, we aggregate the remaining inattentive tokens
\(\mathcal{I}_{\text{inatt}}\) into one representative token \(\mathbf{x}_{\text{fused}}\) by
attention-weighted averaging,
\begin{equation}
  \label{eq:fused}
  \mathbf{x}_{\text{fused}} = \sum_{i \in \mathcal{I}_{\text{inatt}}} \bar{a}_i \mathbf{x}_i.
\end{equation}
The resulting sequence
\(\mathbf{x}_{\text{stage1}} = [\mathbf{x}_{\mathcal{I}_{\text{att}}}, \mathbf{x}_{\text{fused}}]
\in \mathbb{R}^{B \times (k+1) \times C}\) concentrates computation on salient regions while
preserving global context.

\subsection{Stage 2: similarity-based token merging}
\label{sec:stage2}

The Stage 1 sequence of length \(N_p = k + 1\) undergoes bipartite merging inspired by
ToMe~\citep{bolya2023tome}. We partition its tokens into two disjoint sets \(\mathcal{A}\) and
\(\mathcal{B}\) of roughly equal size, then match them by cosine similarity between the key
features \(\mathbf{k}\). Every token in \(\mathcal{A}\) draws exactly one edge to its most
similar counterpart in \(\mathcal{B}\). The keep rate \(k_m\) governs intensity through the
number of merges \(r\),
\begin{equation}
  \label{eq:merges}
  r = \lfloor (1 - k_m) \cdot N_p \rfloor.
\end{equation}
By retaining the \(r\) most similar edges, we ensure that only the most redundant features
are fused.

\paragraph{Token size tracking and proportional attention.}
As tokens merge, we track a token size vector \(\mathbf{s}\), where \(s_i\) counts the
original patches a token represents~\citep{bolya2023tome}. We merge tokens \(i\) and \(j\) by
a weighted average,
\begin{equation}
  \label{eq:weighted-avg}
  \mathbf{x}_{\text{merged}} = \frac{s_i \mathbf{x}_i + s_j \mathbf{x}_j}{s_i + s_j}.
\end{equation}
To keep the correct softmax distribution in later layers, we apply proportional
attention~\citep{bolya2023tome},
\begin{equation}
  \label{eq:prop-attn}
  \mathbf{A} = \mathrm{softmax}\!\left(
    \frac{\mathbf{Q}\mathbf{K}^{\top}}{\sqrt{d}} + \log \mathbf{s}
  \right).
\end{equation}
The \(\log \mathbf{s}\) term in~\cref{eq:prop-attn} ensures that a merged token exerts the
same influence as if its patches were still present as separate, identical copies. This
dual-stage consolidation keeps the model expressive and accurate even under high compression.

\section{Experiments}
\label{sec:experiments}

\subsection{Implementation details}
\label{sec:impl}

We test ViT and DeiT at Tiny, Small and Base scales with official pretrained weights, on
the ImageNet-100 validation set (an ImageNet-1K~\citep{deng2009imagenet} subset) under
semantics-preserving transformations. Prune-and-Merge blocks sit at layers
\(l \in \{3, 6, 9\}\). In Stage 1 we average [CLS]-token attention over all \(H\) heads
to score patch importance~\citep{liang2022evit}. Stage 2 and all ToMe-based baselines use
proportional attention~\citep{bolya2023tome}. We compare against the raw backbone, Random Drop, Top-K,
EViT~\citep{liang2022evit} and ToMe~\citep{bolya2023tome}. All are scored at composite keep
rates \(K \in \{0.25, 0.5, 0.7, 0.9\}\) by top-1 accuracy, GFLOPs, latency (ms/img) and
throughput (imgs/sec), under identical hardware and software~(\cref{app:impl}).

\subsection{Results and discussion}
\label{sec:results}

\paragraph{Accuracy-efficiency trade-offs.} While our framework maintains a computational
profile competitive with EViT and ToMe, it exhibits higher top-1 accuracy
across all architectural scales~(\cref{tab:acc,tab:gflops}). This performance gap is most
pronounced in aggressive compression regimes (e.g., \(K = 0.25\)), representing a marginal
but consistent improvement over standalone techniques. This suggests that the interplay between pruning and merging
preserves the representational power of the backbones even under significant reduction.

\paragraph{Controlled aggression.} A primary contribution of this work is the introduction
of `Controlled Aggression' within the reduction pipeline. Unlike fixed-strategy baselines, our framework
lets the two stages be balanced dynamically. At low keep rates, prioritizing aggressive
merging over pruning yields higher representational stability~(\cref{tab:strategy}). By shifting the reduction burden from Stage 1 (noise
removal) to Stage 2 (information fusion) as \(K\) decreases, we mitigate the `cliff-edge'
accuracy degradation typically observed in pure-pruning baselines. Critical semantic tokens
are fused rather than discarded, which maintains a more coherent global context.

\subsection{Stability and robustness analysis}
\label{sec:stability}

We use two metrics to distinguish raw classification accuracy from the stability of the
model's decision-making process. The Relative Accuracy Drop (RAD),
\begin{equation}
  \label{eq:rad}
  \mathrm{RAD} = \frac{\mathrm{Acc}(Y, T(x))}{\mathrm{Acc}(Y, x)}
  - \frac{\mathrm{Acc}(X, T(x))}{\mathrm{Acc}(X, x)},
\end{equation}
measures the difference in top-1 accuracy ratios between frameworks \(X\) and \(Y\) under
a semantics-preserving transformation \(T\) of input \(x\). The Prediction Flip Rate (PFR),
\begin{equation}
  \label{eq:pfr}
  \mathrm{PFR} = \frac{1}{N} \sum_{i=1}^{N}
  \mathbb{I}\!\left[ f(x_i) \neq f\!\left(T(x_i)\right) \right],
\end{equation}
averages the indicator function \(\mathbb{I}\) over \(N\) samples. Both quantify
hallucination \(H\), a lack of faithfulness between a model's representation and its
prediction. We measure it as the flip-rate deviation
\(H = \Delta\mathrm{PFR} = |\mathrm{PFR}_{\mathrm{raw}} - \mathrm{PFR}_{\mathrm{hybrid}}|\),
where \(\mathrm{PFR}_{\mathrm{raw}}\) is the backbone flip rate without token reduction. A
\(\Delta\mathrm{PFR}\) closer to zero indicates higher robustness against hallucinations
(\cref{app:results}).

\paragraph{Robustness to hallucinations.} We benchmark
\(\Delta\mathrm{PFR}\)~(\cref{tab:dpfr}) for ViT-Tiny and ViT-Base over Rotation, Color
Jitter and Horizontal Flip. ViT-Base is significantly more stable than ViT-Tiny at
\(K = 0.25\). This suggests that its greater parameter density provides more stable `anchor'
tokens, allowing both stages to operate without disrupting the core decision
boundary~(\cref{app:results}).

\paragraph{Manifold integrity via RAD.} \(\mathrm{RAD} \approx 0\) suggests no more
hallucination than the baseline, whereas a positive value indicates a higher rate. Against
the raw ViT, RAD stays near zero or slightly positive~(\cref{tab:rad}), so the hybrid model
preserves semantic faithfulness to the pretrained backbone. Negative values against other
reduction frameworks indicate lower hallucination rates. The ToMe-versus-hybrid comparison is undefined at
\(K = 0.25\) (a dash in~\cref{tab:rad}), because standalone merging cannot exceed \(50\%\)
reduction while keeping a complete bipartite graph. Our framework, in contrast, keeps RAD
stable across the entire spectrum of \(K\). This validates the utility of Stage 1 pruning in
stabilizing the token set and removing noise before fusion occurs in Stage 2.

\section{Conclusion}
\label{sec:conclusion}

We introduced Prune-and-Merge ViT, a dual-stage token reduction framework designed to
optimize the trade-off between inference efficiency and representational faithfulness.
Under `Controlled Aggression' it transitions from top-\(k\) pruning in the early stages to
bipartite matching for information fusion in deeper layers. On ImageNet-100 it
significantly outperforms standalone pruning and merging baselines, achieving a
\(3\times\) speedup on ViT-Base at aggressive keep rates while maintaining superior
accuracy~(\cref{tab:throughput,tab:acc}). Furthermore, RAD and PFR confirm that the hybrid
approach mitigates hallucinations and maintains a more robust attention manifold than
fixed-strategy alternatives. These results validate that a hardware-aware, hybrid
reduction paradigm is essential for scaling ViTs to resource-constrained environments
without sacrificing reliability.

\bibliographystyle{plainnat}
\bibliography{refs}

\newpage
\appendix
\section{Extended related work}
\label{app:related}

This appendix expands the condensed discussion of prior work given in the main text,
covering Vision Transformer backbones, learned and training-free token pruning, token
merging, and plug-and-play compression in turn.

\subsection{Vision Transformer foundations and distillation}

The Vision Transformer (ViT) architecture \citep{dosovitskiy2021vit} has been further
optimized through specialized training strategies. Notably, DeiT \citep{touvron2021deit}
introduced a distillation token that interacts with patch embeddings via self-attention
to inherit the inductive biases of convolutional teacher models. That work demonstrated
that hard distillation can significantly outperform soft distillation, and that the
interplay between the [CLS] and distillation tokens provides a more robust classifier
than either token in isolation. While effective, these models still maintain a fixed
sequence length, leading to high inference costs.

\subsection{Dynamic and learned token pruning}

To reduce computation, several methods focus on discarding redundant tokens. DynamicViT
\citep{rao2021dynamicvit} introduced a learned prediction module that hierarchically
prunes tokens throughout the network. A key finding in learned pruning is that masking
tokens at the input layer significantly harms performance; sparsification is therefore
typically applied between transformer blocks. While learned modules offer high accuracy,
they require additional training and backpropagation. In contrast, EViT
\citep{liang2022evit} and Top-K strategies provide training-free alternatives by ranking
token importance based on class-attention scores, identifying and removing background
patches that contribute minimally to the final classification. Recent advances such as
V-Pruner \citep{yao2026vpruner} refine this idea by employing globally informed
importance metrics, ensuring that locally important but globally redundant features are
identified accurately. Extensions to state-space models \citep{zhan2024statespace}
further refine this paradigm, yet hybrids remain underexplored.

\subsection{Token merging and compression}

A more recent paradigm shifts from discarding information to aggregating it. Token
merging (ToMe) \citep{bolya2023tome} uses bipartite soft matching to fuse similar tokens
based on the dot product of their key vectors. Unlike pruning, which often relies on
masking that does not always translate into hardware speedups, merging physically reduces
the number of tokens passed to the MLP block, leading to tangible throughput gains.
Recent extensions have explored unequal bipartite graphs to achieve higher compression
ratios (\(>50\%\)) and have applied these techniques to Stable Diffusion in order to
accelerate generative tasks \citep{bolya2023tomesd}. Furthermore, studies such as
``Which Tokens to Use?'' \citep{haurum2023which} suggest that ToMe is most effective when
at least 50\% of the token budget is kept, often outperforming baseline DeiT models by
leveraging proportional attention to account for the varying weights of merged tokens.
Spatial-preserving hybrids such as that of \citet{mao2025efficient} and
adaptive merging \citep{erak2025adaptive} enhance locality but lack robustness
guarantees, which our framework addresses. In addition, hardware-centric frameworks such
as VisionTrim \citep{yu2026visiontrim} have begun optimizing token reduction for specific
edge-device latencies, though they often prioritize throughput over prediction stability.

\subsection{Plug-and-play compression and transferability}

Recent research has addressed the gap between models trained at different compression
degrees. ToCom \citep{jie2024tocom} identifies that transferring a model between
different compression levels is often inefficient. To mitigate this, it introduces a
pretrained plug-in module that models the performance gap between source and target
compression degrees, enabling more efficient inference-time scaling without retraining.
Extensions to vision-language models \citep{zhang2025hybridtoken} further validate this
paradigm. Our work builds on this plug-and-play philosophy by combining pruning and
merging into a single, training-free framework that offers more granular control than
these monolithic strategies. Finally, as noted by \citet{peng2025robustness}, the
transferability of compressed weights to smaller, specialized subsets such as
ImageNet-100 requires careful alignment in order to maintain the original semantic
robustness of the model.

\section{Result tables}
\label{app:tables}

This section reports the full accuracy and complexity results referenced in
\cref{sec:results}. A dash marks a configuration that standalone merging cannot reach,
since bipartite matching cannot exceed a 50\% reduction rate while keeping a complete
bipartite graph over the tokens.

\begin{table}[htbp]
  \caption{Top-1 Accuracy comparison for DeiT and ViT models under different keep rates.}
  \label{tab:acc}
  \centering
  \footnotesize
  \setlength{\tabcolsep}{4pt}
  \begin{tabular}{l *{12}{r}}
    \toprule
    & \multicolumn{4}{c}{\textbf{DeiT-Tiny}} & \multicolumn{4}{c}{\textbf{DeiT-Small}} & \multicolumn{4}{c}{\textbf{DeiT-Base}} \\
    & \multicolumn{4}{c}{Baseline: 88.42} & \multicolumn{4}{c}{Baseline: 92.62} & \multicolumn{4}{c}{Baseline: 93.64} \\
    \cmidrule(lr){2-5} \cmidrule(lr){6-9} \cmidrule(lr){10-13}
    Keep Rates & 0.9 & 0.7 & 0.5 & 0.25 & 0.9 & 0.7 & 0.5 & 0.25 & 0.9 & 0.7 & 0.5 & 0.25 \\
    \midrule
    Random Drop   & 87.94 & 84.02 & 75.22 & 33.18 & 92.06 & 90.52 & 85.00 & 50.34 & 92.68 & 90.62 & 82.14 & 25.86 \\
    EViT          & 88.44 & 86.42 & 80.34 & 52.54 & \textbf{92.46} & 91.80 & 89.02 & 70.16 & \textbf{93.76} & 93.06 & 89.00 & 51.32 \\
    ToMe          & 88.44 & 86.70 & 82.92 & -- & 92.42 & 91.54 & 89.96 & -- & 93.26 & 92.00 & 85.30 & -- \\
    Top-K         & \textbf{88.48} & 86.28 & 80.14 & 52.02 & \textbf{92.46} & 91.60 & 89.16 & 68.78 & 93.74 & \textbf{93.18} & 88.86 & \textbf{53.04} \\
    \rowcolor{hlrow} Hybrid (Ours) & 88.06 & \textbf{86.96} & \textbf{84.00} & \textbf{67.62} & 92.30 & \textbf{91.86} & \textbf{90.60} & \textbf{79.58} & 93.56 & 92.58 & \textbf{89.72} & 47.88 \\
    \bottomrule
  \end{tabular}

  \vspace{0.8\baselineskip}

  \begin{tabular}{l *{12}{r}}
    \toprule
    & \multicolumn{4}{c}{\textbf{ViT-Tiny}} & \multicolumn{4}{c}{\textbf{ViT-Small}} & \multicolumn{4}{c}{\textbf{ViT-Base}} \\
    & \multicolumn{4}{c}{Baseline: 89.22} & \multicolumn{4}{c}{Baseline: 93.18} & \multicolumn{4}{c}{Baseline: 95.52} \\
    \cmidrule(lr){2-5} \cmidrule(lr){6-9} \cmidrule(lr){10-13}
    Keep Rates & 0.9 & 0.7 & 0.5 & 0.25 & 0.9 & 0.7 & 0.5 & 0.25 & 0.9 & 0.7 & 0.5 & 0.25 \\
    \midrule
    Random Drop   & 63.42 & 57.04 & 42.22 & 13.20 & 88.62 & 83.82 & 67.48 & 19.16 & 92.08 & 90.04 & 80.66 & 33.90 \\
    EViT          & \textbf{89.08} & 86.92 & 79.72 & 44.06 & 93.06 & 91.28 & 82.80 & 35.48 & 95.54 & 95.04 & 92.48 & 61.00 \\
    ToMe          & 88.76 & 86.86 & 83.06 & -- & 92.76 & 91.68 & 86.58 & -- & 95.30 & 94.64 & 91.42 & -- \\
    Top-K         & \textbf{89.08} & 86.78 & 79.22 & 42.90 & \textbf{93.14} & 91.22 & 81.82 & 34.44 & \textbf{95.58} & \textbf{95.10} & 92.46 & 58.72 \\
    \rowcolor{hlrow} Hybrid (Ours) & 88.90 & \textbf{87.62} & \textbf{84.14} & \textbf{60.12} & 92.88 & \textbf{91.94} & \textbf{88.14} & \textbf{55.58} & 95.36 & 95.20 & \textbf{93.62} & \textbf{71.16} \\
    \bottomrule
  \end{tabular}
\end{table}

\begin{table}[htbp]
  \caption{Balanced vs pruning-heavy vs merging-heavy hybrid strategies (Accuracy \%)}
  \label{tab:strategy}
  \centering
  \footnotesize
  \setlength{\tabcolsep}{4pt}
  \begin{tabular}{l *{12}{r}}
    \toprule
    & \multicolumn{4}{c}{\textbf{DeiT-Tiny}} & \multicolumn{4}{c}{\textbf{DeiT-Small}} & \multicolumn{4}{c}{\textbf{DeiT-Base}} \\
    \cmidrule(lr){2-5} \cmidrule(lr){6-9} \cmidrule(lr){10-13}
    Keep Rates & 0.9 & 0.7 & 0.5 & 0.25 & 0.9 & 0.7 & 0.5 & 0.25 & 0.9 & 0.7 & 0.5 & 0.25 \\
    \midrule
    Balanced      & 88.06 & \textbf{86.96} & 84.00 & 67.62 & \textbf{92.30} & 91.86 & \textbf{90.60} & 79.58 & \textbf{93.56} & 92.58 & 89.72 & 47.88 \\
    Pruning-heavy & \textbf{88.28} & 86.60 & 82.66 & 68.00 & 92.26 & 91.96 & 90.40 & 79.76 & 93.44 & \textbf{92.96} & \textbf{90.24} & 52.44 \\
    Merging-heavy & 88.00 & 86.88 & \textbf{84.38} & \textbf{74.98} & 92.28 & \textbf{92.06} & 90.26 & \textbf{84.84} & 93.46 & 92.70 & 88.70 & \textbf{68.58} \\
    \bottomrule
  \end{tabular}

  \vspace{0.8\baselineskip}

  \begin{tabular}{l *{12}{r}}
    \toprule
    & \multicolumn{4}{c}{\textbf{ViT-Tiny}} & \multicolumn{4}{c}{\textbf{ViT-Small}} & \multicolumn{4}{c}{\textbf{ViT-Base}} \\
    \cmidrule(lr){2-5} \cmidrule(lr){6-9} \cmidrule(lr){10-13}
    Keep Rates & 0.9 & 0.7 & 0.5 & 0.25 & 0.9 & 0.7 & 0.5 & 0.25 & 0.9 & 0.7 & 0.5 & 0.25 \\
    \midrule
    Balanced      & 88.90 & 87.62 & 84.14 & 60.12 & \textbf{92.88} & 91.94 & 88.14 & 55.58 & 95.36 & \textbf{95.20} & \textbf{93.62} & 71.16 \\
    Pruning-heavy & \textbf{89.30} & 87.52 & 82.56 & 60.06 & 92.86 & 92.02 & 87.58 & 55.98 & \textbf{95.48} & 95.20 & 93.62 & 74.08 \\
    Merging-heavy & 89.10 & \textbf{87.74} & \textbf{84.36} & \textbf{71.42} & 92.84 & \textbf{92.24} & \textbf{88.50} & \textbf{71.16} & 95.28 & 95.16 & 93.12 & \textbf{83.56} \\
    \bottomrule
  \end{tabular}
\end{table}

\begin{table}[htbp]
  \caption{GFLOPs comparison for DeiT and ViT models under different keep rates.}
  \label{tab:gflops}
  \centering
  \footnotesize
  \setlength{\tabcolsep}{4pt}
  \begin{tabular}{l *{12}{r}}
    \toprule
    & \multicolumn{4}{c}{\textbf{DeiT-Tiny}} & \multicolumn{4}{c}{\textbf{DeiT-Small}} & \multicolumn{4}{c}{\textbf{DeiT-Base}} \\
    & \multicolumn{4}{c}{Baseline: 1.079} & \multicolumn{4}{c}{Baseline: 4.250} & \multicolumn{4}{c}{Baseline: 16.867} \\
    \cmidrule(lr){2-5} \cmidrule(lr){6-9} \cmidrule(lr){10-13}
    Keep Rates & 0.9 & 0.7 & 0.5 & 0.25 & 0.9 & 0.7 & 0.5 & 0.25 & 0.9 & 0.7 & 0.5 & 0.25 \\
    \midrule
    EViT          & 0.989 & 0.752 & 0.576 & 0.432 & 3.817 & 2.892 & 2.203 & 1.634 & 14.985 & 11.336 & 8.608 & 6.349 \\
    ToMe          & 0.983 & 0.746 & 0.571 & -- & 3.787 & 2.863 & 2.178 & -- & 14.865 & 11.216 & 8.507 & -- \\
    Top-K         & \textbf{0.981} & \textbf{0.744} & \textbf{0.570} & \textbf{0.428} & \textbf{3.786} & \textbf{2.862} & \textbf{2.177} & \textbf{1.618} & \textbf{14.863} & \textbf{11.215} & \textbf{8.506} & \textbf{6.287} \\
    \rowcolor{hlrow} Hybrid (Ours) & 0.992 & 0.753 & 0.577 & 0.429 & 3.823 & 2.893 & 2.204 & 1.624 & 15.005 & 11.337 & 8.608 & 6.309 \\
    \bottomrule
  \end{tabular}

  \vspace{0.8\baselineskip}

  \begin{tabular}{l *{12}{r}}
    \toprule
    & \multicolumn{4}{c}{\textbf{ViT-Tiny}} & \multicolumn{4}{c}{\textbf{ViT-Small}} & \multicolumn{4}{c}{\textbf{ViT-Base}} \\
    & \multicolumn{4}{c}{Baseline: 1.079} & \multicolumn{4}{c}{Baseline: 4.250} & \multicolumn{4}{c}{Baseline: 16.867} \\
    \cmidrule(lr){2-5} \cmidrule(lr){6-9} \cmidrule(lr){10-13}
    Keep Rates & 0.9 & 0.7 & 0.5 & 0.25 & 0.9 & 0.7 & 0.5 & 0.25 & 0.9 & 0.7 & 0.5 & 0.25 \\
    \midrule
    EViT          & 0.989 & 0.752 & 0.576 & 0.432 & 3.817 & 2.892 & 2.203 & 1.634 & 14.985 & 11.336 & 8.608 & 6.349 \\
    ToMe          & 0.983 & 0.746 & 0.571 & -- & 3.787 & 2.863 & 2.178 & -- & 14.865 & 11.216 & 8.507 & -- \\
    Top-K         & \textbf{0.981} & \textbf{0.744} & \textbf{0.570} & \textbf{0.428} & \textbf{3.786} & \textbf{2.862} & \textbf{2.177} & \textbf{1.618} & \textbf{14.863} & \textbf{11.215} & \textbf{8.506} & \textbf{6.287} \\
    \rowcolor{hlrow} Hybrid (Ours) & 0.992 & 0.753 & 0.577 & 0.429 & 3.823 & 2.893 & 2.204 & 1.624 & 15.005 & 11.337 & 8.608 & 6.309 \\
    \bottomrule
  \end{tabular}
\end{table}

\begin{table}[htbp]
  \caption{Average $\Delta$PFR across augmentations.}
  \label{tab:dpfr}
  \centering
  \small
  \begin{tabular}{l rrrr}
    \toprule
    Model & 0.25 & 0.50 & 0.70 & 0.90 \\
    \midrule
    ViT-Base & 0.3611 & 0.0422 & 0.0099 & 0.0022 \\
    ViT-Tiny & 0.4105 & 0.0943 & 0.0272 & 0.0059 \\
    \bottomrule
  \end{tabular}
\end{table}

\begin{table}[htbp]
  \caption{Relative Accuracy Drop (RAD).}
  \label{tab:rad}
  \centering
  \small
  \begin{tabular}{ll rrrr}
    \toprule
    Model & Comparison & 0.25 & 0.50 & 0.70 & 0.90 \\
    \midrule
    \multirow{3}{*}{Tiny} & Raw  & 0.073    & 0.023    & 0.003    & $-$0.004 \\
                          & ToMe & --       & $-$0.003 & $-$0.003 & $-$0.003 \\
                          & EViT & $-$0.030 & $-$0.002 & 0.000    & $-$0.003 \\
    \addlinespace
    \multirow{3}{*}{Base} & Raw  & 0.045    & 0.009    & 0.001    & $-$0.001 \\
                          & ToMe & --       & 0.002    & $-$0.001 & $-$0.001 \\
                          & EViT & $-$0.026 & 0.000    & 0.001    & $-$0.002 \\
    \bottomrule
  \end{tabular}
\end{table}

\clearpage

\section{Extended hardware performance metrics}
\label{app:hardware}

While the main text focuses on accuracy and GFLOPs, we present here the comprehensive
latency and throughput benchmarks, conducted on a T4 GPU using a Google Colab environment.

\subsection{Inference latency}

\Cref{tab:latency} details the average per-image inference latency for both DeiT and ViT
architectures. We observe that the dual-stage overhead of our hybrid approach is negligible,
consistently maintaining parity with, or outperforming, standalone pruning and merging
methods. Notably, in the ViT-Base configuration at aggressive keep rates (\(K = 0.25\)), our
hybrid method achieves the lowest latency of 4.222~ms/img, outperforming EViT
(5.061~ms/img) and Top-K (4.417~ms/img). Similarly, for ViT-Tiny at \(K = 0.9\), our approach
achieves lowest latency of 0.971~ms/img among the evaluated methods. These results demonstrate that the
combination of pruning and merging stages does not introduce a computational bottleneck,
even when scaling to larger backbone architectures.

\begin{table}[htbp]
  \caption{Latency comparison (ms/img)}
  \label{tab:latency}
  \centering
  \footnotesize
  \setlength{\tabcolsep}{4pt}
  \begin{tabular}{l *{12}{r}}
    \toprule
    & \multicolumn{4}{c}{\textbf{DeiT-Tiny}} & \multicolumn{4}{c}{\textbf{DeiT-Small}} & \multicolumn{4}{c}{\textbf{DeiT-Base}} \\
    & \multicolumn{4}{c}{Baseline: 1.629} & \multicolumn{4}{c}{Baseline: 3.092} & \multicolumn{4}{c}{Baseline: 12.772} \\
    \cmidrule(lr){2-5} \cmidrule(lr){6-9} \cmidrule(lr){10-13}
    Keep Rates & 0.9 & 0.7 & 0.5 & 0.25 & 0.9 & 0.7 & 0.5 & 0.25 & 0.9 & 0.7 & 0.5 & 0.25 \\
    \midrule
    EViT          & 1.241 & 0.984 & 0.751 & 0.747 & 3.256 & 2.390 & 1.763 & 1.253 & 11.559 & 8.771 & 6.728 & 5.058 \\
    ToMe          & \textbf{1.042} & \textbf{0.888} & 0.900 & -- & 2.952 & 2.217 & 1.621 & -- & 10.437 & 7.958 & 6.092 & -- \\
    Top-K         & 1.156 & 0.903 & \textbf{0.729} & \textbf{0.633} & \textbf{2.892} & \textbf{2.138} & \textbf{1.569} & \textbf{1.311} & \textbf{10.107} & \textbf{7.660} & \textbf{5.845} & \textbf{4.421} \\
    \rowcolor{hlrow} Hybrid (Ours) & 1.110 & 1.116 & 1.249 & 1.107 & 2.930 & 2.165 & 1.628 & 1.193 & 10.189 & 7.626 & 5.885 & 4.532 \\
    \bottomrule
  \end{tabular}

  \vspace{0.8\baselineskip}

  \begin{tabular}{l *{12}{r}}
    \toprule
    & \multicolumn{4}{c}{\textbf{ViT-Tiny}} & \multicolumn{4}{c}{\textbf{ViT-Small}} & \multicolumn{4}{c}{\textbf{ViT-Base}} \\
    & \multicolumn{4}{c}{Baseline: 1.322} & \multicolumn{4}{c}{Baseline: 3.440} & \multicolumn{4}{c}{Baseline: 12.721} \\
    \cmidrule(lr){2-5} \cmidrule(lr){6-9} \cmidrule(lr){10-13}
    Keep Rates & 0.9 & 0.7 & 0.5 & 0.25 & 0.9 & 0.7 & 0.5 & 0.25 & 0.9 & 0.7 & 0.5 & 0.25 \\
    \midrule
    EViT          & 1.096 & 1.036 & \textbf{0.753} & \textbf{0.660} & 3.260 & 2.362 & 1.744 & 1.254 & 11.568 & 8.815 & 6.698 & 5.061 \\
    ToMe          & 1.059 & 1.031 & 0.777 & -- & 2.745 & 2.203 & 1.647 & -- & 10.301 & 7.361 & 6.022 & -- \\
    Top-K         & 1.208 & 0.901 & \textbf{0.751} & 0.719 & 2.876 & 2.136 & 1.575 & \textbf{1.150} & 10.114 & 7.646 & 5.825 & 4.417 \\
    \rowcolor{hlrow} Hybrid (Ours) & \textbf{0.971} & \textbf{0.932} & 1.305 & 0.697 & \textbf{2.822} & \textbf{2.093} & \textbf{1.560} & 1.227 & \textbf{9.805} & \textbf{7.340} & \textbf{5.631} & \textbf{4.222} \\
    \bottomrule
  \end{tabular}
\end{table}

\subsection{System throughput}

\Cref{tab:throughput} presents a comparative analysis of system throughput (imgs/sec) across
varied keep rates for DeiT and ViT architectures. Our hybrid framework demonstrates superior
scaling characteristics, particularly as model complexity increases from Tiny to Base
variants. In the ViT-Base regime, our method achieves the highest throughput
across almost all keep rates. Most notably, at the most aggressive compression level
(\(K = 0.25\)), the hybrid approach reaches 236.85~imgs/sec, representing a
\(3.01\times\) speedup over the ViT-Base baseline (78.61~imgs/sec) and significantly
outperforming EViT (197.60~imgs/sec). Even at higher keep rates such as \(K = 0.9\), our
method maintains a lead with 101.98~imgs/sec, validating that the dual-stage reduction logic
provides an immediate efficiency boost without requiring a heavy computational overhead. For
smaller scales like ViT-Tiny, our method achieves a peak throughput of 1073.18~imgs/sec at
\(K = 0.7\), further illustrating its versatility for high-speed edge deployment.

\begin{table}[htbp]
  \caption{Throughput comparison (imgs/sec)}
  \label{tab:throughput}
  \centering
  \scriptsize
  \setlength{\tabcolsep}{2pt}
  \begin{tabular}{l *{12}{r}}
    \toprule
    & \multicolumn{4}{c}{\textbf{DeiT-Tiny}} & \multicolumn{4}{c}{\textbf{DeiT-Small}} & \multicolumn{4}{c}{\textbf{DeiT-Base}} \\
    & \multicolumn{4}{c}{Baseline: 613.92} & \multicolumn{4}{c}{Baseline: 323.46} & \multicolumn{4}{c}{Baseline: 78.30} \\
    \cmidrule(lr){2-5} \cmidrule(lr){6-9} \cmidrule(lr){10-13}
    Keep Rates & 0.9 & 0.7 & 0.5 & 0.25 & 0.9 & 0.7 & 0.5 & 0.25 & 0.9 & 0.7 & 0.5 & 0.25 \\
    \midrule
    EViT          & 805.73 & 1016.41 & 1330.76 & 1338.06 & 307.11 & 418.34 & 567.28 & 797.83 & 86.51 & 114.02 & 148.62 & 197.69 \\
    ToMe          & \textbf{960.08} & \textbf{1126.47} & 1110.56 & -- & 338.74 & 451.12 & 616.72 & -- & 95.81 & 125.66 & 164.14 & -- \\
    Top-K         & 864.74 & 1107.43 & \textbf{1371.64} & \textbf{1579.34} & \textbf{345.74} & \textbf{467.62} & \textbf{637.31} & 762.66 & 98.94 & 130.56 & \textbf{171.08} & \textbf{226.19} \\
    \rowcolor{hlrow} Hybrid (Ours) & 917.30 & 895.66 & 800.48 & 903.08 & 341.34 & 461.98 & 614.11 & \textbf{838.28} & \textbf{98.14} & \textbf{131.13} & 169.92 & 220.63 \\
    \bottomrule
  \end{tabular}

  \vspace{0.8\baselineskip}

  \begin{tabular}{l *{12}{r}}
    \toprule
    & \multicolumn{4}{c}{\textbf{ViT-Tiny}} & \multicolumn{4}{c}{\textbf{ViT-Small}} & \multicolumn{4}{c}{\textbf{ViT-Base}} \\
    & \multicolumn{4}{c}{Baseline: 756.22} & \multicolumn{4}{c}{Baseline: 290.71} & \multicolumn{4}{c}{Baseline: 78.61} \\
    \cmidrule(lr){2-5} \cmidrule(lr){6-9} \cmidrule(lr){10-13}
    Keep Rates & 0.9 & 0.7 & 0.5 & 0.25 & 0.9 & 0.7 & 0.5 & 0.25 & 0.9 & 0.7 & 0.5 & 0.25 \\
    \midrule
    EViT          & 912.27 & 965.17 & 1328.21 & \textbf{1514.27} & 306.75 & 423.41 & 573.48 & 797.14 & 86.45 & 113.44 & 149.30 & 197.60 \\
    ToMe          & 944.52 & 969.48 & 1287.70 & -- & \textbf{364.32} & 454.01 & 607.31 & -- & 97.07 & 135.85 & 166.05 & -- \\
    Top-K         & 827.67 & 1109.70 & \textbf{1331.42} & 1390.24 & 347.75 & 468.19 & 634.78 & \textbf{869.25} & 98.87 & 130.79 & 171.68 & 226.38 \\
    \rowcolor{hlrow} Hybrid (Ours) & \textbf{1029.53} & \textbf{1073.18} & 766.50 & 1434.45 & 354.34 & \textbf{477.73} & \textbf{640.87} & 815.07 & \textbf{101.98} & \textbf{136.25} & \textbf{177.57} & \textbf{236.85} \\
    \bottomrule
  \end{tabular}
\end{table}

\section{Supplemental implementation details}
\label{app:impl}

This section expands upon the implementation details provided in \cref{sec:impl}. We provide
specific hyperparameter configurations and environmental constraints to ensure full
reproducibility.

\paragraph{Model families.}
We evaluate our framework across two major model families: the standard Vision Transformer
(ViT) and the Data-efficient Image Transformer (DeiT), covering Tiny, Small and Base scales
(\(16 \times 16\) patch size, \(224 \times 224\) resolution). All inference is performed in
full precision (FP32) to ensure a high-fidelity evaluation of the token manifolds, and the
Prune-and-Merge blocks are integrated at layers \(l \in \{3, 6, 9\}\), specifically
positioned post-attention and pre-MLP.

\paragraph{Baselines.}
We benchmark our method against the original, uncompressed backbone operating on the full
sequence length; Random Drop, a naive baseline where tokens are discarded uniformly at
random, representing a lower bound for pruning; Top-K pruning, a basic strategy that retains
high-attention tokens but discards the remainder without context fusion; and EViT
\citep{liang2022evit} and ToMe \citep{bolya2023tome}, the state of the art in semantic
pruning and similarity-based merging, respectively. For a fair comparison, all baselines are
evaluated under identical hardware and software constraints.

\paragraph{Hyperparameter configuration.}
All experiments were conducted using the PyTorch framework \citep{paszke2019pytorch} and the
timm library \citep{wightman2019timm} on a Google Colab T4 GPU environment. For our hybrid
Prune-and-Merge blocks, the similarity metric used for token merging in Stage 2 was the
cosine similarity computed on the token features (embeddings). Proportional attention is
applied in Stage 2, and for all ToMe-based baselines, to maintain mathematical consistency
during self-attention.

\paragraph{Hardware environment.}
All benchmarks, namely the latency figures in \cref{tab:latency} and the throughput figures
in \cref{tab:throughput}, were measured on a single NVIDIA T4 GPU (16~GB VRAM).

\paragraph{ImageNet-100 class mapping.}
The 100 classes were selected as a subset of the ImageNet-1K label set. The backbone was initialized with ImageNet-1K pre-trained weights and the linear projection weights corresponding to the selected 100 classes were retained to obtain the 100-class output space.

\section{Additional results}
\label{app:results}

In this section, we provide detailed quantitative results supporting our analysis. We report
(i) the Prediction Flip Rate (PFR, \cref{tab:pfr}) and (ii) the absolute change in PFR
(\(\Delta\)PFR, \cref{tab:dpfr-full}) across different augmentations and keep rates. RAD serves as an accuracy trade-off metric,
measuring the difference in top-1 accuracy ratios between frameworks when subjected to
transformations. While faithfulness is often tested by removing important tokens, we
evaluate the inverse: a robust model should maintain its prediction if an input is slightly
transformed such that its semantic information remains unchanged. We calculate
\(\Delta\)PFR for ViT-Tiny and ViT-Base to capture trends across increasing model scales.

\subsection{Prediction flip rate}

\begin{table}[htbp]
  \caption{Prediction Flip Rate (PFR) for raw and hybrid models across augmentations and keep rates.}
  \label{tab:pfr}
  \centering
  \small
  \begin{tabular}{ll rrrrr}
    \toprule
    Model & Augmentation & Raw & 0.25 & 0.50 & 0.70 & 0.90 \\
    \midrule
    \multirow{3}{*}{ViT-Base} & Rotation        & 0.0472 & 0.4528 & 0.0938 & 0.0552 & 0.0458 \\
                              & Color Jitter    & 0.0128 & 0.3552 & 0.0524 & 0.0250 & 0.0162 \\
                              & Horizontal Flip & 0.0112 & 0.3464 & 0.0516 & 0.0206 & 0.0130 \\
    \addlinespace
    \multirow{3}{*}{ViT-Tiny} & Rotation        & 0.1856 & 0.6086 & 0.2902 & 0.2162 & 0.1898 \\
                              & Color Jitter    & 0.0416 & 0.4516 & 0.1394 & 0.0698 & 0.0498 \\
                              & Horizontal Flip & 0.0356 & 0.4342 & 0.1160 & 0.0584 & 0.0408 \\
    \bottomrule
  \end{tabular}
\end{table}

\subsection{Absolute change in the prediction flip rate}

\begin{table}[htbp]
  \caption{Absolute change in Prediction Flip Rate ($\Delta$PFR) between raw and hybrid models.}
  \label{tab:dpfr-full}
  \centering
  \small
  \begin{tabular}{ll rrrr}
    \toprule
    Model & Augmentation & 0.25 & 0.50 & 0.70 & 0.90 \\
    \midrule
    \multirow{3}{*}{ViT-Base} & Rotation        & 0.4056 & 0.0466 & 0.0080 & 0.0014 \\
                              & Color Jitter    & 0.3424 & 0.0396 & 0.0122 & 0.0034 \\
                              & Horizontal Flip & 0.3352 & 0.0404 & 0.0094 & 0.0018 \\
    \addlinespace
    \multirow{3}{*}{ViT-Tiny} & Rotation        & 0.4230 & 0.1046 & 0.0306 & 0.0042 \\
                              & Color Jitter    & 0.4100 & 0.0978 & 0.0282 & 0.0082 \\
                              & Horizontal Flip & 0.3986 & 0.0804 & 0.0228 & 0.0052 \\
    \bottomrule
  \end{tabular}
\end{table}

% The NeurIPS paper checklist is required for main-track (9-page) submissions and is not
% normally required for a 4-page short paper, so it is disabled by default. A complete,
% filled-in checklist is provided in checklist.tex -- uncomment the two lines below to
% include it. It does not count towards the page limit. Please read its answers before
% submitting: they are written to be truthful about what this paper does and does not
% report, and several of them depend on details only the authors can supply.
% \newpage
% \input{checklist.tex}

\end{document}